%% file: arxiv.tex
\documentclass[10pt,twocolumn,letterpaper]{article}
\usepackage[pagenumbers]{cvpr}
\input{preamble}
\definecolor{cvprblue}{rgb}{0.21,0.49,0.74}
\usepackage[pagebackref,breaklinks,colorlinks,allcolors=cvprblue]{hyperref}
\def\confName{CVPR}
\def\confYear{2027}
\title{HOPD: Historical On-Policy Distillation for SAR MLLM Continual Learning}
\author{
Yuxuan Li$^{1}$ \quad Derrick Goh Xin Deik$^{2}$ \quad Zheng-Peng Duan$^{3}$\\
Qianru Sun$^{1}$ \quad Weijie Li$^{4}\dagger$\\[3pt]
{\normalsize $^{1}$Singapore Management University \quad $^{2}$Nanyang Technological University}\\
{\normalsize $^{3}$Nankai University \quad $^{4}$Xi'an Jiaotong University}
}
\input{review/supplement-review-macros}
\hypersetup{pdftitle={HOPD: Historical On-Policy Distillation for SAR MLLM Continual Learning},
 pdfauthor={Yuxuan Li, Derrick Goh Xin Deik, Zheng-Peng Duan, Qianru Sun, Weijie Li}}
\begin{document}
\maketitle
\begin{abstract}
Continual SAR interpretation requires multimodal large language models to learn incoming sources while retaining earlier task abilities.
This is difficult when acquisition and geographic shifts alter visual evidence, regionally concentrated categories bias answers, and historical images are scarce.
We first formulate and then address this problem with Historical On-Policy Distillation (HOPD), which reuses source-acquisition LoRA states as persistent teachers for one evolving student.
Archiving each state before consolidation preserves a stable source reference.
The teachers supervise student-generated prefixes, transferring historical behavior to the student's changing generation contexts.
To enrich sparse replay, original-view teachers guide augmented historical student views while preserving spatial coordinates.
This knowledge is consolidated into a single deployed adapter that requires no source identity.
On SARVQA-CL, our source-incremental evaluation protocol, HOPD achieves the highest final mean score of 54.51 among the compared methods, exceeding the strongest compared method by 2.96 points.
It also improves the final source-macro score in every evaluated task over shared-memory Replay-SFT, with gains across source orders.
These results support source-acquisition history as supervision for retaining earlier task performance during continual SAR adaptation.
Code and data: https://github.com/zcablii/HOPD.
\end{abstract}

\input{sec/1_intro}
\input{sec/2_related_work}
\input{sec/3_problem_observations}
\input{sec/4_method}
\input{sec/5_experiments}
\input{sec/6_conclusion}

{\small
\bibliographystyle{ieeenat_fullname}
\bibliography{main}}
\clearpage
\appendix
\setcounter{figure}{0}
\setcounter{table}{0}
\setcounter{equation}{0}
\renewcommand{\theHfigure}{supp.\arabic{figure}}
\renewcommand{\theHtable}{supp.\arabic{table}}
\renewcommand{\theHequation}{supp.\arabic{equation}}
\setcounter{dbltopnumber}{3}
\twocolumn[{\centering\Large\bfseries Supplementary Material\par\vspace{1em}}]
\input{sec/S_supplement}

\end{document}

%% file: preamble.tex
\usepackage{microtype}
\usepackage{tabularx}
\usepackage{multirow}
\usepackage{array}
\usepackage{algorithm}
\usepackage{algpseudocode}
\usepackage{tikz}
\usetikzlibrary{arrows.meta,positioning,fit,calc,backgrounds}
\definecolor{methodblue}{HTML}{246AA6}
\definecolor{memoryorange}{HTML}{B66A24}
\definecolor{softblue}{HTML}{EAF2F8}
\definecolor{softorange}{HTML}{FCF2E7}
\newcommand{\method}{HOPD\xspace}
\newcommand{\dataset}{SARVQA-CL\xspace}

\newcommand{\NA}{\textemdash}
\newcommand{\E}{\mathbb{E}}
\newcommand{\KL}{\mathrm{KL}}
\newcommand{\sg}{\operatorname{sg}}
\newcommand{\vocab}{\mathcal{V}}
\newcommand{\memory}{\mathcal{M}}
\newcommand{\metric}[1]{\mathrm{#1}}
\newcolumntype{Y}{>{\raggedright\arraybackslash}X}
\newcommand{\sartile}[2]{\begin{tikzpicture}
\clip (0,0) rectangle (#2,#2);
\pgfmathsetmacro{\atlaswidth}{3*#2}
\node[anchor=south west,inner sep=0] at ({-#1*#2},0)
 {\includegraphics[width=\atlaswidth cm]{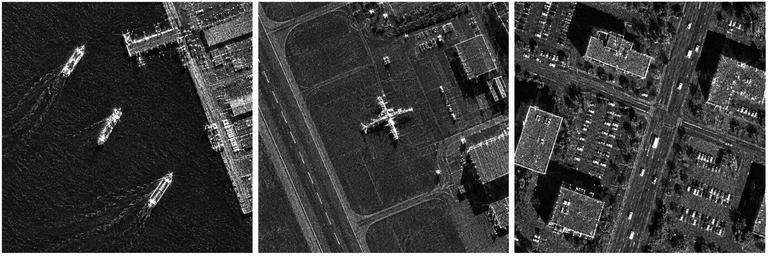}};
\end{tikzpicture}}

\newcommand{\sargammatile}[2]{\begin{tikzpicture}
\clip (0,0) rectangle (#2,#2);
\pgfmathsetmacro{\atlaswidth}{3*#2}
\node[anchor=south west,inner sep=0] at ({-#1*#2},0)
 {\includegraphics[width=\atlaswidth cm]{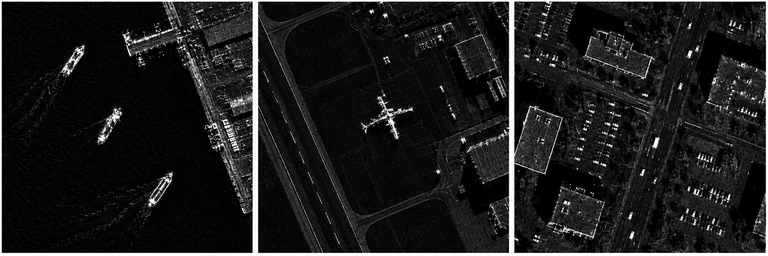}};
\end{tikzpicture}}

%% file: review/supplement-review-macros.tex
\definecolor{suppreviewred}{RGB}{185,20,28}
\ifdefined\SupplementReviewEnabled
  \ifdefined\XeTeXversion\else
    \PackageError{supplement-review}{Use XeLaTeX for Chinese review notes}{Compile supplement-review.tex with xelatex.}
  \fi
  \usepackage{fontspec}
  \newfontfamily\suppreviewfont{Microsoft YaHei}
\fi
\DeclareRobustCommand{\reviewmark}[1]{\ifdefined\SupplementReviewEnabled{\color{suppreviewred}#1}\else#1\fi}

%% file: sec/1_intro.tex
\input{fig/motivation}
\section{Introduction}
\label{sec:intro}
Synthetic aperture radar (SAR) supports all-weather, day-and-night observation for maritime surveillance, infrastructure monitoring, and other interpretation services~\cite{moreira2013tutorial,ma2025sarchat}.
As sensors and regional collections expand, interpretation models must incorporate newly annotated imagery while remaining reliable on earlier sources.
Continual learning~\cite{parisi2019continual,delange2021survey} addresses this maintenance requirement by learning from incoming data while preserving acquired capabilities~\cite{kirkpatrick2017ewc,lopezpaz2017gem}.
For multimodal large language models (MLLMs), the challenge is to acquire new SAR visual knowledge while retaining earlier task abilities.

Prior continual-learning settings examine different forms of increment.
General MLLM benchmarks introduce abilities such as OCR, mathematical reasoning, and GUI interaction~\cite{zhao2025mllmcl}, while SAR studies address target aspects or class increments~\cite{chen2025aspect,li2025sarcil}.
CLeaRS extends continual vision-language evaluation to remote-sensing task and modality transitions~\cite{weng2026clears}.
We study successive sources within SAR under a common task interface, where the model continues to answer what objects are present, how many there are, and where they are located.
We formulate \emph{source-incremental continual SAR VQA}, with limited historical data access and no source identity at inference, and establish \dataset by adapting SARFact~\cite{wang2026sarfact} to this protocol.

Although the visual questions remain stable, the visual evidence and answer statistics change across sources (\cref{fig:motivation}).
Radar frequency, polarization, and viewing geometry affect scattering~\cite{moreira2013tutorial}, while regional collections differ in land cover and target categories~\cite{wei2026sarlang}.
At the same time, ship- or aircraft-focused sources repeatedly expose the model to concentrated categories~\cite{li2024sardet100k}.
These shifts motivate examining answer-prior bias~\cite{agrawal2018priors} alongside SAR appearance variation and background cues~\cite{dong2024background}.
With sparse historical images, retaining earlier task abilities requires a stable reference for the interpretation behavior acquired from each source.

Our insight is that the learning process itself provides these references.
The LoRA state~\cite{hu2022lora} acquired from an incoming source captures its interpretation behavior before subsequent updates can overwrite it.
We turn these states into persistent source-associated teachers in \emph{Historical On-Policy Distillation} (HOPD).
Each teacher is archived before consolidation and revisited through a small set of retained image--question pairs.
This makes the model's source-acquisition history available as supervision throughout later updates.

Transferring this historical knowledge must account for the student's changing generation contexts and the scarcity of replay images.
HOPD queries each historical teacher on the student's own generated prefixes~\cite{agarwal2024gkd}, providing soft targets at contexts encountered by the updated model.
Following cross-view visual distillation~\cite{li2026s2vopd}, the student receives a gamma-transformed historical image while the teacher evaluates the original image and the same prefix.
This asymmetry diversifies visual conditioning while preserving spatial coordinates.
Current-source learning and deployment use original images, and consolidation produces a single adapter without source routing.

On \dataset, HOPD improves the final mean score over the strongest compared baseline by 2.96 points.
It also outperforms Replay-SFT by 7.45 points under the same image memory and similar acquisition performance.
The gains persist across source orders and all task-level source averages.
Shared-endpoint controls measure recovery during consolidation, while behavioral diagnostics track the preservation of earlier category and spatial predictions.
Together, these results support source-acquisition history as a reference for retaining knowledge during repeated SAR updates.

Our contributions are threefold:
\begin{itemize}
 \item To our knowledge, we provide the first dedicated study of source-incremental continual learning for SAR MLLMs, with the established \dataset to evaluate source-incremental continual SAR VQA tasks.
 \item We propose HOPD, which turns source-acquisition LoRA states into persistent supervision. Student-prefix distillation and asymmetric historical views transfer these source references into one evolving adapter while diversifying sparse image replay.
 \item HOPD delivers the highest final mean performance among the compared methods. Results across source orders and behavioral diagnostics show sustained retention of category and spatial facts.
\end{itemize}

%% file: fig/motivation.tex
\begin{figure}[t]
\centering
\begin{tikzpicture}[x=1cm,y=1cm,font=\scriptsize,>=Latex]
\definecolor{retain}{RGB}{0,120,94}
\definecolor{forget}{RGB}{190,48,44}
\path[use as bounding box] (0,.12) rectangle (8.2,-8.00);
\node[anchor=west,font=\footnotesize\bfseries] at (.05,-.06) {(a) New abilities vs. new visual sources};
\node[text=black!65] at (1.89,-.46) {Ability-incremental MLLM CL};
\node[text=methodblue,font=\scriptsize\bfseries] at (6.10,-.46) {Continual SAR interpretation};
\draw[black!20] (3.93,-.69)--(3.93,-2.04);
\node[draw=black!30,fill=black!3,rounded corners=2pt,minimum width=1.25cm,minimum height=.84cm,font=\large\bfseries] at (.85,-1.19) {A7};
\node[draw=black!30,fill=black!3,rounded corners=2pt,minimum width=1.25cm,minimum height=.84cm,font=\large] at (2.97,-1.19) {$2+3$};
\draw[->,thick,black!55] (1.56,-1.19)--(2.24,-1.19);
\node at (.85,-1.88) {Read text};
\node at (2.97,-1.88) {Solve math};
\foreach \idx/\xx in {0/4.72,2/7.38}{
 \node[inner sep=0] at (\xx,-1.25) {\sartile{\idx}{1.03}};
}
\node[inner sep=0] at (6.05,-1.25) {\includegraphics[width=1.03cm,height=1.03cm]{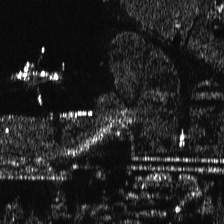}};
\draw[->,methodblue,thick] (5.28,-1.25)--(5.48,-1.25);
\draw[->,methodblue,thick] (6.61,-1.25)--(6.81,-1.25);
\node[text=methodblue] at (6.05,-1.98) {Same questions: what? how many? where?};
\draw[black!18] (.05,-2.25)--(8.15,-2.25);
\node[anchor=west,font=\footnotesize\bfseries] at (.05,-2.55) {(b) Acquisition + geographic shifts};
\node[text=black!65] at (2.00,-2.94) {Different acquisition conditions};
\node[text=black!65] at (6.13,-2.94) {Same sensor, different regions};
\node[inner sep=0] at (1.02,-3.80) {\sartile{0}{1.35}};
\node[inner sep=0] at (2.98,-3.80) {\sargammatile{0}{1.35}};
\draw[<->,methodblue,thick] (1.77,-3.80)--(2.23,-3.80);
\node[inner sep=0] at (5.15,-3.80) {\sartile{0}{1.35}};
\node[inner sep=0] at (7.11,-3.80) {\sartile{2}{1.35}};
\draw[<->,methodblue,thick] (5.90,-3.80)--(6.36,-3.80);
\node at (2.00,-4.72) {Scattering / appearance};
\node at (6.13,-4.72) {Land cover / target populations};
\draw[black!18] (.05,-4.99)--(8.15,-4.99);
\node[anchor=west,font=\footnotesize\bfseries] at (.05,-5.29) {(c) Regional category bias can overwrite old facts};
\node[text=black!65] at (1.02,-5.73) {Incoming maritime data};
\node[text=retain] at (4.22,-5.73) {Old image: before};
\node[text=forget] at (6.98,-5.73) {Same image: after};
\node[inner sep=0] at (1.02,-6.65) {\sartile{0}{1.50}};
\node[inner sep=0] at (4.22,-6.65) {\includegraphics[width=1.50cm,height=1.50cm]{fig/assets/sadd_02215_aircraft.png}};
\node[inner sep=0] at (6.98,-6.65) {\includegraphics[width=1.50cm,height=1.50cm]{fig/assets/sadd_02215_aircraft.png}};
\draw[->,memoryorange,line width=1pt] (1.87,-6.65)--node[above,align=center,text=forget,font=\scriptsize\bfseries]{ship\\ship\\ship}(3.25,-6.65);
\draw[->,forget,thick] (5.15,-6.65)--(6.04,-6.65);
\draw[retain,line width=1.1pt] (3.537,-6.630) rectangle (3.925,-6.188);
\draw[white,dashed,line width=.9pt] (6.297,-6.630) rectangle (6.685,-6.188);
\draw[forget,line width=1.1pt] (7.348,-6.951) rectangle (7.563,-6.697);
\node[fill=forget,text=white,inner sep=1pt,font=\tiny] at (7.455,-6.655) {ship};
\node[text=black!65] at (1.02,-7.66) {Dominant answer: ship};
\node[text=retain,font=\scriptsize\bfseries] at (4.22,-7.66) {aircraft + correct box};
\node[text=black,align=center,font=\scriptsize\bfseries] at (6.98,-7.66) {\textcolor{forget}{wrong ship}\\and missing aircraft};
\end{tikzpicture}
\caption{\textbf{Continual SAR interpretation under evolving visual sources.} (a) Stable interpretation tasks require adaptation to new visual evidence. (b) Acquisition and geographic shifts alter SAR appearance. (c) Regionally concentrated categories can overwrite earlier recognition and grounding.}
\label{fig:motivation}
\end{figure}

%% file: sec/2_related_work.tex
\section{Related Work}
\label{sec:related_work}

\subsection{Vision-Language Learning for SAR}
SAR representation learning provides a complementary foundation for interpretation.
At the backbone level, previous works advance remote-sensing visual representations through context-aware learning~\cite{li2025lsknet,saratrxv2}, multimodal learning~\cite{li2026sm3det,sarvia}, and language-guided pretraining~\cite{li2025visual,pmlr-v306-li26q}.
Gradient-based joint-embedding prediction~\cite{li2024predicting} and SARATR-X~\cite{li2025saratr} develop self-supervised representations for target recognition.
ATRNet-STAR~\cite{liu2026atrnet} broadens recognition benchmarks with diverse targets and realistic imaging conditions.

Vision-language learning builds language-guided interpretation through image--text pretraining~\cite{radford2021clip,wang2026slip,wang2026winardet,li2023blip2} and visual instruction tuning~\cite{liu2023llava,dai2023instructblip}.
In remote sensing, these capabilities support question answering~\cite{lobry2020rsvqa}, transferable image--text representations~\cite{liu2024remoteclip}, and grounded conversation~\cite{kuckreja2024geochat}.
Recent SAR work develops visual-language supervision tailored to radar imagery.
SARChat-Bench-2M and SARLANG-1M expand this foundation through dialogue and image--text corpora~\cite{ma2025sarchat,wei2026sarlang}, while SARVLM studies domain-transfer training~\cite{ma2026sarvlm}.
Retrieval~\cite{ramirez2026sarrag} and optical fusion~\cite{tosato2025sarstrikes} provide complementary information for SAR VQA.
These approaches improve visual representations, cross-modal alignment, and language-guided interpretation from available collections.
However, HOPD addresses their continued maintenance within SAR: preserving earlier task abilities while a single deployed model learns successive visual sources.

\subsection{Continual Adaptation of MLLMs}
Continual learning addresses the tension between acquiring new capabilities and retaining earlier knowledge.
For MLLMs, this involves both instruction alignment and factual competence, whose distinct forgetting behaviors are examined in CoIN and SEFE~\cite{chen2024coin,chen2025sefe}.
Existing retention strategies revisit stored examples or logits~\cite{rebuffi2017icarl,chaudhry2019tiny,buzzega2020der,aljundi2019mir}, constrain gradients or important parameters~\cite{chaudhry2019agem,kirkpatrick2017ewc,zenke2017si,aljundi2018mas}, or restrict adapter updates as in RegLoRA~\cite{chen2025sefe}.
An alternative allocates prompts or adapter capacity across updates through learned prompts~\cite{wang2022l2p,wang2022dualprompt}, orthogonal subspaces~\cite{wang2023olora}, or routed adapters~\cite{zhao2025mllmcl}.
In remote sensing, CLeaRS studies task and modality transitions~\cite{weng2026clears}, while SAR studies address target-aspect and class increments~\cite{chen2025aspect,li2025sarcil}.
To our knowledge, we present the first dedicated study of source-incremental continual learning for SAR MLLMs.

\subsection{On-Policy Knowledge Distillation}
Distillation preserves behavior through teacher predictions~\cite{gou2021kdsurvey,li2016lwf}, including sequence-level targets for autoregressive models~\cite{kim2016seqkd}.
Its effectiveness also depends on where those predictions are queried.
Learning on learner-induced contexts addresses sequential prediction shift~\cite{ross2011dagger}, and GKD applies this principle by distilling on student rollouts~\cite{agarwal2024gkd}.
Within on-policy distillation, MiniLLM studies reverse-KL matching~\cite{gu2024minillm}, while SDFT obtains teacher supervision through demonstration conditioning~\cite{shenfeld2026sdft}.
Nemotron-Cascade 2 consolidates capability domains using their best validation checkpoints from Cascade RL and sampled-token reverse-KL advantages~\cite{yang2026nemotron}.
HOPD instead freezes each source's LoRA after acquisition and before consolidation, selecting teachers by replay-query provenance.
Full-vocabulary forward KL on student prefixes transfers these historical references into one adapter under limited image memory.

\subsection{Cross-View Visual Distillation}
Complementary visual conditioning can make teacher predictions informative beyond matching the student's input.
Earlier consistency learning and self-distillation use moving-average teachers to supervise perturbed views~\cite{tarvainen2017meanteacher,caron2021dino}.
Vision-OPD transfers crop-conditioned perception to full-image students~\cite{yuan2026visionopd}.
S$^2$VOPD matches original-view teacher predictions to augmented student rollouts using an EMA teacher and Jensen--Shannon divergence, and also evaluates a frozen base-model teacher~\cite{li2026s2vopd}.
Other approaches exploit content-erased contrasts~\cite{liang2026vcsd} or sensitivity to visual detail~\cite{liu2026vaopd}.
While these methods improve perception through visual conditioning, HOPD couples cross-view supervision with source-acquisition teachers to preserve knowledge through later updates.
Gamma transformations diversify historical student views without changing localization coordinates, while current-source inputs remain unchanged.

%% file: sec/3_problem_observations.tex
\section{Problem Setting}
\label{sec:problem}
We consider annotated SAR sources $\mathcal D_1,\ldots,\mathcal D_T$ arriving sequentially to update an existing interpretation model.
An example $(I,q,y,k)$ contains an image, a question, an answer, and a task type $k\in\{1,\ldots,K\}$.
The objective is to learn each new source while preserving these capabilities on previously observed sources.

At stage $t$, training can access the current source $\mathcal D_t$ and the declared historical memory and adapters.
Full training data from completed sources are unavailable to subsequent updates.
After each stage, evaluation covers all seen sources using the same task interface.
The question identifies the requested task, but inference receives no source identifier.
\Cref{sec:experiments} specifies the source and task inventory used to instantiate this general formulation.

%% file: sec/4_method.tex
\section{Historical On-Policy Distillation}
\label{sec:method}
\input{fig/method}
We propose Historical On-Policy Distillation (\method), which preserves source-specific interpretation knowledge in one continually updated MLLM by consolidating stable source-acquisition references on student-generated contexts and augmented historical views.
Let $\theta$ denote the trainable LoRA parameters, with the backbone fixed, and $\theta_{t-1}$ their state before source $t$.
Stage A produces the adapted state $\theta_t^A$, archived as frozen teacher $\phi_t$. Stage B consolidates source-associated teachers into $\theta_t$, which serves inference and initializes the next source.

\subsection{Source-acquisition teachers}
\label{sec:adaptation}
Stage A learns the current source $t$ through supervised fine-tuning initialized from $\theta_{t-1}$:
\begin{equation}
 \mathcal L_A(\theta)=\E_{(x,y)\sim\mathcal D_t}
 \left[-\frac{1}{|y|}\sum_{\ell=1}^{|y|}
 \log p_\theta(y_\ell\mid x,y_{<\ell})\right],
 \label{eq:sft}
\end{equation}
Here, $\mathcal L_A$ is the supervised loss and $\E$ denotes expectation over sampled QA.
The input $x=(I,q)$ pairs image $I$ with task-formatted question $q$, and $p_\theta$ is the model's conditional token distribution.
For answer sequence $y$, $|y|$ is its length, $y_\ell$ its $\ell$th token, and $y_{<\ell}$ the preceding prefix.
Tasks are sampled uniformly, and the loss averages tokens within each QA.
Adaptation uses no historical replay.

Freezing $\phi_t\leftarrow\theta_t^A$ immediately after adaptation preserves a source reference without training a separate expert.
Unlike the latest student, which may have lost older behavior, the fixed archive $\Phi_t=\{\phi_1,\ldots,\phi_t\}$ retains each source's acquisition-time reference.

\paragraph{Shared-image replay.}
\label{sec:memory}
For each completed source indexed by $i$, memory $\memory_i$ retains $n$ training images, where $n$ is a small per-source image budget.
Each image contributes one native QA for each of the $K$ tasks, giving $nK$ QA per source.
Images are selected randomly.
During source $t$'s consolidation, historical queries come from memory and current-source queries use the full training data:
\begin{equation}
 \begin{split}
 i&\sim\mathrm{Unif}\{1,\ldots,t\},\quad
 k\sim\mathrm{Unif}\{1,\ldots,K\},\\
 (I,q)&\sim
 \begin{cases}
  \memory_i^{(k)}, & i<t,\\
  \mathcal D_t^{(k)}, & i=t.
 \end{cases}
 \end{split}
 \label{eq:query_sampling}
\end{equation}
Here, $k$ indexes tasks, $\mathrm{Unif}$ denotes uniform sampling, and superscript $(k)$ selects task-$k$ queries from memory or current-source data.
Each historical pool $\memory_i^{(k)}$ contains $n$ image--query pairs. Examples come from shuffled per-task pools.
We commit $\memory_t$ after consolidation, or after adaptation at the first source, where consolidation is skipped.

\subsection{Historical replay views}
\label{sec:gamma}
To diversify sparse historical inputs, we apply gamma transformations to the student view while keeping each frozen teacher on the original image~\cite{li2026s2vopd}:
\begin{align}
 G_\gamma(I)&=\operatorname{round}\!\left[255(I/255)^\gamma\right],
 \quad \gamma\sim\mathcal U[0.6,1.6),\label{eq:gamma}\\
 I^s_{i,t}&=
 \begin{cases}
  G_\gamma(I), & i<t,\\
  I, & i=t.
 \end{cases}\label{eq:student_view}
\end{align}
Here, $G_\gamma$ is an elementwise intensity transform with exponent $\gamma$ drawn uniformly from $[0.6,1.6)$.
Image values lie in $[0,255]$, and $\operatorname{round}$ rounds to the nearest integer.
$I^s_{i,t}$ denotes the student view of a source-$i$ image at stage $t$, with $s$ indicating the student.
The shared RGB transform preserves pixel locations and coordinate targets.
The original teacher view supplies a stable target, while current-source queries remain untransformed.

\subsection{Student-prefix distillation}
\label{sec:opd}
Following on-policy distillation~\cite{agarwal2024gkd}, we sample one response per selected QA using the student view:
\begin{equation}
 \widehat y\sim p_{\theta^-}(\cdot\mid I^s_{i,t},q),
 \label{eq:rollout}
\end{equation}
Here, $\widehat y$ is the sampled token sequence and $\theta^-$ is the adapter at the start of the update.
Sampling uses temperature 1 without top-$k$/nucleus truncation and stops at a stop token or task cap.
We reuse the \emph{same transformed image and integer token sequence} for the student loss, while the teacher evaluates those prefixes on the original image:
\begin{align}
 p_\ell&=p_\theta(\cdot\mid I^s_{i,t},q,\widehat y_{<\ell}),\nonumber\\
 h_\ell&=\sg\!\left[p_{\phi_i}(\cdot\mid I,q,\widehat y_{<\ell})\right].
 \label{eq:contexts}
\end{align}
Here, $p_\ell$ and $h_\ell$ are student and teacher token distributions at position $\ell$, conditioned on the sampled prefix $\widehat y_{<\ell}$.
The operator $\sg$ stops gradients through its argument.
The Stage-B loss $\mathcal L_B$ averages the forward-KL divergence $d(p,h)=\KL(h\|p)$ from teacher $h$ to student $p$:
\begin{align}
 \mathcal L_B(\theta)&=\E_{i,k,(I,q),\gamma,\widehat y}
  \left[\frac1L\sum_{\ell=1}^L d(p_\ell,h_\ell)\right],
  \label{eq:opd}\\
 d(p,h)&=\sum_{v\in\vocab}h(v)\log\frac{h(v)}{p(v)}.
 \label{eq:fkl}
\end{align}
Here, $\vocab$ is the token vocabulary, $v$ indexes its entries, and $L=|\widehat y|$ includes a sampled stop token when present.
The expectation follows \cref{eq:query_sampling,eq:gamma,eq:rollout}.
All vocabulary entries contribute, gradients act on student log probabilities, and QA receive equal weight.
Generated sequences remain fixed during differentiation, so distribution matching uses no policy gradients through sampling.

\paragraph{Why supervise student prefixes?}
CE replay uses stored hard answers on annotated prefixes, while fixed-prefix KD supplies teacher soft targets~\cite{hinton2015distilling} on prescribed prefixes.
HOPD instead queries acquisition-time teachers on student prefixes, addressing GKD's training--generation mismatch~\cite{agarwal2024gkd} without adding historical images. Retained SAR queries thus provide supervision for changing predictions rather than a repeated fixed answer trajectory.

\subsection{Recursive consolidation}
\label{sec:resources}
The consolidated $\theta_t$ initializes the next source.
Inference uses one adapter and original images without source routing, perturbation, or teacher ensembles, and the trainable parameter count stays fixed.
The supplement gives the step-by-step execution algorithm.

%% file: fig/method.tex
\begin{figure*}[t]
\centering
\newcommand{\hopdfrozenicon}{\tikz[baseline=-.55ex,x=1cm,y=1cm,sharp corners]{
 \foreach \a in {0,60,...,300} {
  \begin{scope}[rotate=\a]
   \draw[methodblue,line width=.45pt,line cap=round,sharp corners]
    (0,0)--(0,.16) (-.045,.13)--(0,.085)--(.045,.13);
  \end{scope}
 }
}}
\newcommand{\hopdtrainicon}{\tikz[baseline=-.55ex,x=1cm,y=1cm,sharp corners]{
 \path[fill=red!85!black,draw=none,sharp corners]
  (0,-.15)
  .. controls (-.13,-.15) and (-.16,-.055) .. (-.12,.025)
  .. controls (-.10,.065) and (-.075,.085) .. (-.075,.115)
  .. controls (-.075,.045) and (-.025,.035) .. (-.025,.08)
  .. controls (-.025,.13) and (.045,.17) .. (.025,.23)
  .. controls (.10,.17) and (.15,.065) .. (.14,-.025)
  .. controls (.14,-.105) and (.075,-.15) .. (0,-.15) -- cycle;
 \path[fill=orange!60!yellow,draw=none,sharp corners]
  (0,-.13)
  .. controls (-.07,-.13) and (-.08,-.075) .. (-.045,-.03)
  .. controls (-.02,.005) and (.02,.025) .. (.02,.07)
  .. controls (.08,-.01) and (.105,-.11) .. (0,-.13) -- cycle;
}}

\begin{tikzpicture}[x=1cm,y=1cm,font=\scriptsize,
 box/.style={draw=black!25,fill=white,rounded corners=3pt,align=center,inner sep=5pt},
 flow/.style={-{Latex[length=1.6mm,width=1.2mm]},line width=.65pt,draw=methodblue,shorten <=.3mm,shorten >=.3mm},
 history/.style={flow,draw=memoryorange}]
\path[use as bounding box] (0,.15) rectangle (17.1,-5.74);
\fill[black!4,rounded corners=5pt] (0,.12) rectangle (17.1,-1.22);
\fill[softorange,rounded corners=5pt] (0,-1.5) rectangle (4.1,-5.70);
\fill[softblue,rounded corners=5pt] (4.4,-1.5) rectangle (17.1,-5.70);
\node[align=center] (prev) at (.9,-.53) {Shared student\\$\theta_{t-1}$};
\node[box,text width=3.0cm] (a) at (4.0,-.53) {\textbf{A: current-source SFT}\\Full current TRAIN $\mathcal D_t$};
\node[align=center] (ae) at (7.0,-.53) {Acquired state\\$\theta_t^A$};
\node[box,fill=softblue,text width=3.8cm] (b) at (11.0,-.53) {\textbf{B: historical consolidation}\\Student-prefix distillation ($t>1$)};
\node[align=center] (out) at (15.45,-.53) {\textbf{Deploy $\theta_t$}\\One adapter, no source ID};
\draw[flow] (prev.east)--(a.west);
\draw[flow] (a.east)--(ae.west);
\draw[flow] (ae.east)--(b.west);
\draw[flow] (b.east)--(out.west);
\node[anchor=west,font=\scriptsize\bfseries,text=memoryorange] at (.18,-1.77) {Archive BEFORE consolidation};
\node[box,draw=memoryorange,text width=2.95cm] (archive) at (2.13,-2.43) {$\phi_t\leftarrow\theta_t^A$ (frozen)\\Teacher bank: $\phi_1,\ldots,\phi_t$};
\draw[history] (ae.south)--(7,-1.37)--(.18,-1.37)--(.18,-2.43)--(archive.west);
\node[inner sep=0] at (.80,-3.35) {\sartile{0}{.82}};
\node[inner sep=0] at (1.96,-3.35) {\includegraphics[width=.82cm,height=.82cm]{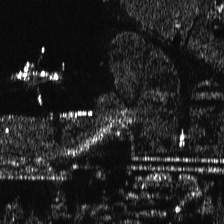}};
\node[inner sep=0] at (3.12,-3.35) {\sartile{2}{.82}};
\node[align=center,text width=3.7cm] (query) at (2.05,-4.65) {\textbf{Sample source $i$ + task uniformly}\\Old: sample retained image + task QA\\Current: full $\mathcal D_t$\\Query provenance selects $\phi_i$};
\node[anchor=west,font=\scriptsize\bfseries,text=methodblue] at (4.6,-1.77) {B: match distributions on the student's own generated context};
\node[inner sep=0,outer sep=0] (orig) at (5.65,-2.80) {\sartile{0}{1.02}};
\node[align=center] at (5.65,-3.46) {Original $I$};
\node[box,draw=memoryorange,text width=2.65cm] (teacher) at (9.05,-2.80) {\mbox{\hopdfrozenicon\hspace{2pt}\textbf{Frozen teacher $\phi_i$}}\\Read student prefix $\widehat y_{<j}$};
\node[align=center] (pt) at (12.35,-2.80) {Teacher target (detached)\\$p_{\phi_i}(\cdot\mid I,q,\widehat y_{<j})$};
\node[inner sep=0,outer sep=0] (view) at (5.65,-4.45) {\sargammatile{0}{1.02}};
\node[align=center] at (5.65,-5.22) {Old: $G_\gamma(I)$\\Current: $I$};
\node[box,draw=methodblue,text width=2.65cm] (student) at (9.05,-4.45) {\mbox{\hopdtrainicon\hspace{2pt}\textbf{Updating student $\theta$}}\\Sample answer $\widehat y$};
\node[align=center] (ps) at (12.35,-4.45) {Student distribution\\$p_\theta(\cdot\mid I^s_{i,t},q,\widehat y_{<j})$};
\node[box,text width=1.6cm] (loss) at (15.65,-3.60) {\textbf{Forward KL}\\$\KL(p_{\phi_i}\|p_\theta)$};
\draw[flow] (orig.east)--(teacher.west);
\draw[flow] (teacher.east)--(pt.west);
\draw[flow] (view.east)--(student.west);
\draw[flow] (student.east)--(ps.west);
\draw[history] (pt.east)-|(loss.north);
\draw[flow] (ps.east)--(14.30,-4.45)--(14.30,-3.72)--([yshift=-3pt]loss.west);
\draw[flow,dashed] (student.north)--node[midway,right,xshift=1.5mm,fill=none,inner sep=0pt]{Sampled prefix $\widehat y_{<j}$}(teacher.south);
\draw[history] (archive.east)--(4.25,-2.43)--(4.25,-2.11)--(9.05,-2.11)--(teacher.north);
\coordinate (inputs) at (4.65,-3.67);
\draw[line width=.65pt,draw=methodblue] (query.east)--(4.28,-4.65)--(4.28,-3.67)--(inputs);
\fill[methodblue] (inputs) circle (.03);
\draw[flow] (inputs)--(4.65,-2.80)--(orig.west);
\draw[flow] (inputs)--(4.65,-4.45)--(view.west);
\draw[flow,line width=.85pt] (loss.south)--(15.65,-5.58)--node[above,inner sep=2pt]{Gradient updates student only}(9.05,-5.58)--(student.south);
\end{tikzpicture}
\caption{\textbf{Historical on-policy distillation.} Stage A archives a source-acquisition teacher before Stage B consolidates seen-source behavior on student-generated prefixes. Teachers receive original images, while historical student images undergo gamma transformation. Only the student receives gradients. Deployment uses one adapter without source identity.}
\label{fig:method}
\end{figure*}

%% file: sec/5_experiments.tex
\section{Experiments}
\label{sec:experiments}
\input{tables/main_results}

\subsection{Experimental setup}
\paragraph{Data and protocol.}
We establish \dataset from SARFact~\cite{wang2026sarfact}, a recent high-quality SAR VQA dataset by merging and post-processing several existing datasets~\cite{zhang2025rsar,li2024sardet100k,wei2020hrsid,wang2026m4}. We only use the nine constituent datasets: OGSOD~\cite{wang2023ogsod}, Shipdataset~\cite{wang2019shipdataset}, SSDD~\cite{zhang2021ssdd}, HRSID~\cite{wei2020hrsid}, SAR-AIRcraft~\cite{wang2023saraircraft}, MSAR~\cite{xia2022msar}, AIR\_SARShip~\cite{sun2019airsarship}, SADD~\cite{zhang2022sadd}, and SIVED~\cite{lin2023sived}.
\dataset contains 356,624 training QA and 5,988 evaluation QA over 2,953 images.
Each source supports the same seven tasks: captioning, counting, presence, classification, detection, referring expressions, and spatial relations.
The supplement details split construction, per-source statistics, and scoring.

\paragraph{Training and repetitions.}
We use Qwen3.5-4B~\cite{qwen2026qwen35} with rank-$8$ LoRA~\cite{hu2022lora} ($\alpha=8$) on the vision encoder, merger, and language model, giving 19.50M trainable parameters.
Stage A uses 1,024 updates per source at learning rate $2\times10^{-4}$, with cosine decay~\cite{loshchilov2017sgdr} and 5\% warmup.
Stage B follows D2--D9 with 128 updates at constant $5\times10^{-5}$.
Both stages use AdamW~\cite{loshchilov2019adamw} and global batch size 8.
By default, Replay-SFT, MIR, MR-LoRA, and HOPD share one retained image and seven task QA per source, totaling nine images and 63 QA after D9.
Stage B follows the source and query sampling in \cref{eq:query_sampling}.
Each method is run three times, once for each of three source orders shared across methods, and results are averaged over these runs.
Training and sampling use seed 42, including memory-image selection by default.
The memory-selection sensitivity analysis changes only the image-selection seed to 43 or 44.
Error bars report variation across orders.
Evaluation uses greedy decoding on original images.
The backbone uses BF16, adapters FP32, and image area is capped at 65,536 pixels with aspect ratio preserved.

\input{fig/current_trajectory}
\paragraph{Baselines.}
We compare sequential LoRA~\cite{hu2022lora}, EWC-LoRA~\cite{kirkpatrick2017ewc}, generative LwF~\cite{li2016lwf}, O-LoRA~\cite{wang2023olora}, RegLoRA from SEFE~\cite{chen2025sefe}, adapted SDFT~\cite{shenfeld2026sdft}, Replay-SFT~\cite{chaudhry2019tiny}, MIR~\cite{aljundi2019mir}, and MR-LoRA~\cite{zhao2025mllmcl}.
Zero-shot evaluates the original Qwen3.5-4B without SAR adaptation.
Sequential LoRA updates one adapter with supervised fine-tuning on the current source alone.
Replay-SFT interleaves current-source QA with historical QA from the shared-image memory under the same objective.

\subsection{Metrics}
\paragraph{Task scores.}
We follow the SARFact evaluation protocol for each task. To be specific, captioning uses annotation-grounded Core Fact F1 over extracted category-presence and explicit count facts.
Detection uses a class-aware adaptation of GRIT~\cite{gupta2022grit} with one-to-one IoU matching~\cite{kuhn1955hungarian}.
Count, presence, category-set classification, and spatial relation use normalized exact accuracy.
Referring uses single-box accuracy at IoU $\geq0.5$~\cite{gupta2022grit}.

\paragraph{Continual learning metrics.}
Let $R_{t,i}$ denote the task-macro score on source $i$ after stage $t$, including B where enabled, with $T=9$.
We summarize acquisition with MFT, final performance with MFN, and performance across stages with MAA~\cite{zhao2025mllmcl,weng2026clears}:
\begin{align}
 \metric{MFT}&=\frac1T\sum_i R_{i,i},\quad
 \metric{MFN}=\frac1T\sum_i R_{T,i},\label{eq:mft}\\
 \metric{MAA}&=\frac1T\sum_t\frac1t\sum_{i\leq t}R_{t,i},\nonumber\\
 \metric{BWT}&=\frac1{T-1}\sum_{i<T}(R_{i,i}-R_{T,i}).\label{eq:cl}
\end{align}
BWT measures performance \emph{drop}, the negative of conventional signed backward transfer~\cite{lopezpaz2017gem}, so lower is better.

\paragraph{SAR behavioral diagnostics.}
We propose four additional metrics to analyze behavioral changes.
To examine what is retained beyond aggregate scores, DCIR measures dominant-class intrusion into previously correct old-source classifications whose reference excludes the arriving source's dominant class.
NFR measures new failures on that set, adapting negative-flip analysis~\cite{yan2021positive}.
JFR tracks retention of previously jointly correct count--detection facts.
JFA measures joint accuracy over all valid fixed pairs, requiring complete category-consistent box matching at IoU $\geq0.5$.
These rates average over old sources, transitions, and orders.
The supplement provides further details.

\subsection{Continual learning performance}
\paragraph{Comparison with state of the art.}
On the \dataset benchmark, HOPD achieves the  best MFN, MAA, and BWT, alongside the best results on all four SAR behavioral diagnostics (\cref{tab:main}).
This combination supports acquiring new-source knowledge while retaining earlier capabilities in a single adapter. The MFN surpasses the second-best method by 2.96 pp, and the outstanding MAA shows that the benefit also holds across the sequence.
Adapted SDFT's weaker acquisition suggests that student-prefix supervision also depends on the teacher.
Its demonstration-conditioned targets~\cite{shenfeld2026sdft} may retain visual or category biases, motivating HOPD's source-acquisition references.

\paragraph{Retention with shared memory.}
With the same image memory, HOPD and Replay-SFT attain similar acquisition performance on MFT, yet HOPD preserves more of it through later updates.
This retention advantage holds in every tested order (\cref{fig:currenttrajectory}), explaining the stronger final performance.
Since acquisition is comparable, the comparison points to better preservation through later updates as the main source of the gain.

\paragraph{Sensitivity to historical image access.}
HOPD uses sparse images effectively: with one image per source, its MFN is close to Replay-SFT with four (\cref{fig:currentmemory}).
Additional images further improve HOPD, showing that its consolidation complements a larger replay memory.
Thus, historical-teacher supervision improves the use of a small image archive while continuing to benefit from broader visual coverage.
Forward KL performs better at the two smaller budgets, supporting its use in the targeted sparse-memory regime.

\input{fig/current_memory}
\input{fig/current_memory_selection_seeds}
\paragraph{Sensitivity to memory-image selection.}
The retention gap changes little when only the stored images are resampled, with acquisition remaining similar between methods (\cref{fig:memoryselectionseeds}).
HOPD's advantage therefore persists beyond the default exemplar selection.
This matters when each source contributes a single image, because the retained example determines which historical scene remains directly available for consolidation.

\input{fig/current_retention_horizon}
\subsection{Retention over subsequent updates}
\paragraph{Aligning sources by retention horizon.}
For source order $o$, we measure forgetting after $k$ subsequent source updates:
\begin{equation}
 F_o(k)=\frac{1}{T-k}\sum_{i=1}^{T-k}
 \left(R^{(o)}_{i,i}-R^{(o)}_{i+k,i}\right).
 \label{eq:retentionhorizon}
\end{equation}
Here, $R$ averages seven task scores, and lower $F_o(k)$ indicates better retention.
Both methods are compared on the same source--stage pairs after equally many updates, with $3(9-k)$ pairs across three orders.

\paragraph{A persistent retention advantage.}
HOPD reduces forgetting at every horizon in every tested order (\cref{fig:retentionhorizon}).
The advantage persists as further sources intervene, rather than appearing only at the final checkpoint.
Across all matched old-source--stage pairs, the average performance drop is reduced by 24.7\% relative to Replay-SFT.

\paragraph{Forgetting from the Stage-A endpoint.}
We also measure forgetting from the Stage-A endpoint, before consolidation can change each source's starting score.
HOPD retains its advantage at every horizon and in every order (Supplementary Table~5).
Both analyses use matched source--stage pairs within each horizon.

\input{tables/mechanism_controls}
\subsection{Consolidation analysis}
\Cref{tab:controls} separates local consolidation effects from complete training outcomes.
Local branches start from common Stage-A-end checkpoints at stages 4 and 9 of order 1, sharing memory with fresh optimizers.
Changes use \emph{No consolidation} as the reference, whereas overall MFN follows each variant's full training trajectory.

\paragraph{Historical recovery.}
HOPD recovers the most old-source performance at both boundaries and better preserves the current source than supervised consolidation.
Old-source recovery benefits from source-acquisition teachers, student prefixes, and asymmetric gamma views relative to their respective ablation experiments.
The prefix comparison keeps historical teachers and image views fixed, directly testing whether supervision should follow the student's generated contexts or annotated answers.
The no-gamma and previous-stage-teacher variants however preserve slightly more current-source performance at stage 9. But the overall MFN indicates HOPD's combination of historical recovery and current-source retention is the most effective for final performance.

\paragraph{Full-sequence gains.}
Across complete training sequences, HOPD outperforms every ablated variant.
The comparisons support student-prefix supervision over supervised consolidation and GT-prefix KD, together with persistent source references and asymmetric replay.
Their benefits therefore extend beyond local recovery to stronger final performance.

\input{fig/current_intrusion}
\subsection{Category and spatial-fact retention}
We test whether HOPD resists incoming category biases and preserves jointly correct counts and locations.

\paragraph{Resistance to category intrusion.}
We evaluate HOPD and Replay-SFT on the same 128 OGSOD classification QA after OGSOD adaptation, before consolidation, and at sequence end across three orders (\cref{fig:currentintrusion}).
Both methods achieve similar initial accuracy, but HOPD retains substantially more recognition ability.
For the illustrated bridge, both initially answer correctly, yet only HOPD retains this answer in every order. Replay-SFT switches to the final source's foreground category.
Across source transitions, HOPD also lowers DCIR and NFR on eligible previously correct old-source answers (\cref{tab:main}), extending the finding beyond this example.

\paragraph{Joint retention of counts and locations.}
We examine count--detection pairs across source transitions to assess whether object quantities and locations remain jointly correct.
HOPD achieves the highest JFR and JFA among all evaluated continual-learning methods, including Replay-SFT, MIR, and MR-LoRA (\cref{tab:main}).
Its lead in both metrics combines stronger preservation of previously correct facts with higher joint accuracy across valid pairs.
These results extend HOPD's retention advantage from category recognition to spatially grounded object facts.

%% file: tables/main_results.tex
\begin{table*}[t]
\centering
\caption{\textbf{Performance comparison of methods in source-incremental SAR VQA on SARVQA-CL.} Results are averaged over three source orders, best values in bold. HOPD achieves the best MFN, MAA, and BWT, alongside the best results on all four SAR behavioral diagnostics.}
\label{tab:main}
\small
\setlength{\tabcolsep}{3pt}
\begin{tabular*}{\textwidth}{@{\extracolsep{\fill}}lcrrrrrrrr@{}}
\toprule
Method & Tunable Params. (M) & MFN$\uparrow$ & MAA$\uparrow$ & BWT$\downarrow$ & MFT$\uparrow$ & DCIR$\downarrow$ & NFR$\downarrow$ & JFR$\uparrow$ & JFA$\uparrow$ \\
\midrule
Zero-shot~\cite{qwen2026qwen35} & 0 & 20.50 & \NA & \NA & \NA & \NA & \NA & \NA & \NA \\
Sequential LoRA~\cite{hu2022lora} & 19.50 & 33.09 & 44.83 & 43.47 & $\boldsymbol{71.73}$ & 84.98 & 91.06 & 20.68 & 9.31 \\
Replay-SFT~\cite{chaudhry2019tiny} & 19.50 & 47.06 & 50.23 & 26.99 & 71.05 & 49.98 & 55.29 & 47.98 & 15.43 \\
EWC-LoRA~\cite{kirkpatrick2017ewc} & 19.50 & 33.74 & 45.22 & 42.36 & 71.39 & 82.97 & 90.00 & 22.49 & 9.90 \\
LwF~\cite{li2016lwf} & 19.50 & 33.01 & 45.73 & 42.96 & 71.19 & 89.36 & 94.92 & 29.60 & 9.73 \\
O-LoRA~\cite{wang2023olora} & 175.47 & 35.95 & 46.01 & 37.38 & 69.18 & 77.88 & 83.39 & 36.51 & 12.10 \\
RegLoRA~\cite{chen2025sefe} & 19.50 & 35.96 & 46.43 & 39.02 & 70.65 & 81.30 & 89.23 & 38.50 & 11.82 \\
SDFT~\cite{shenfeld2026sdft} & 19.50 & 18.66 & 30.11 & 22.18 & 38.38 & 46.44 & 59.71 & 0.00 & 0.00 \\
MIR~\cite{aljundi2019mir} & 19.50 & 50.72 & 51.44 & 19.35 & 67.92 & 38.36 & 46.21 & 45.35 & 14.31 \\
MR-LoRA~\cite{zhao2025mllmcl} & 194.97 & 51.55 & 52.26 & 21.57 & 70.72 & 35.36 & 43.08 & 47.03 & 16.16 \\
\midrule
\textbf{HOPD} (ours) & 19.50 & $\boldsymbol{54.51}$ & $\boldsymbol{54.60}$ & $\boldsymbol{18.58}$ & 71.03 & $\boldsymbol{32.97}$ & $\boldsymbol{39.23}$ & $\boldsymbol{56.06}$ & $\boldsymbol{17.79}$ \\
\bottomrule
\end{tabular*}
\end{table*}

%% file: fig/current_trajectory.tex
\begin{figure*}[t]
\centering
\includegraphics[width=\textwidth]{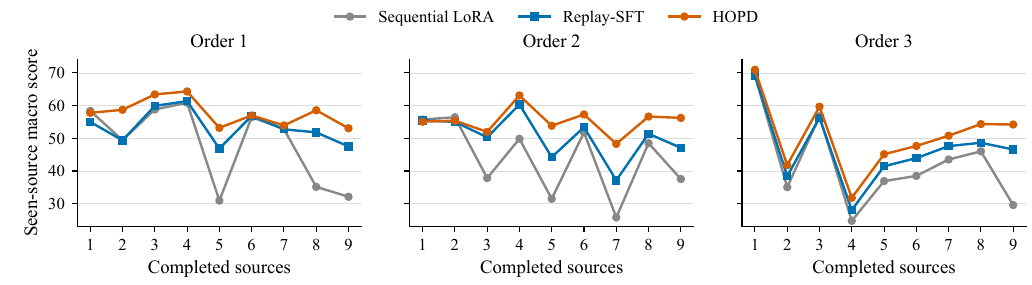}
\vspace{-6pt}
\caption{\textbf{Seen-source performance across source orders.} Each stage averages scores over seen sources and tasks. HOPD finishes above Replay-SFT and Sequential LoRA in every tested order.}
\label{fig:currenttrajectory}
\end{figure*}

%% file: fig/current_memory.tex
\begin{figure}[t]
\centering
\includegraphics[width=\linewidth]{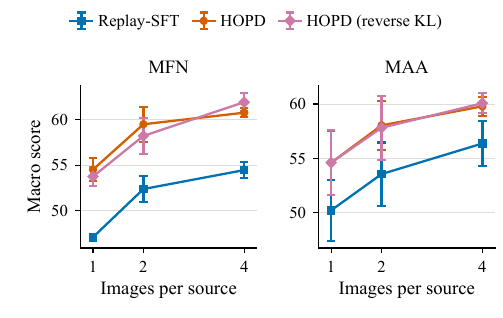}
\vspace{-6pt}
\caption{\textbf{Historical image access and KL direction.} Error bars show source-order SD. HOPD uses forward KL unless marked otherwise. One-image HOPD and four-image Replay-SFT obtain similar final mean scores.}
\label{fig:currentmemory}
\end{figure}

%% file: fig/current_memory_selection_seeds.tex
\begin{figure}[t]
\centering
\includegraphics[width=\linewidth]{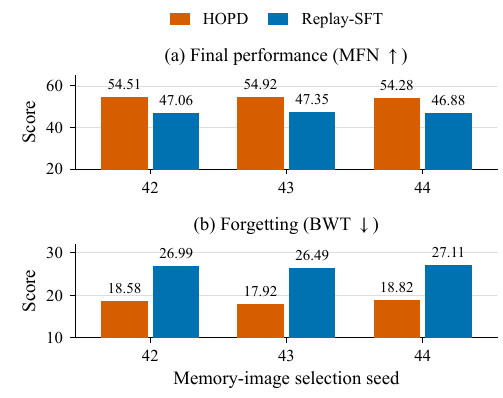}
\caption{\textbf{Memory-selection sensitivity.} MFN and BWT for memory-image selection seeds 42, 43 and 44. Each bar averages three source orders with one retained image per source and all other settings fixed.}
\label{fig:memoryselectionseeds}
\end{figure}

%% file: fig/current_retention_horizon.tex
\begin{figure}[t]
\centering
\includegraphics[width=\linewidth]{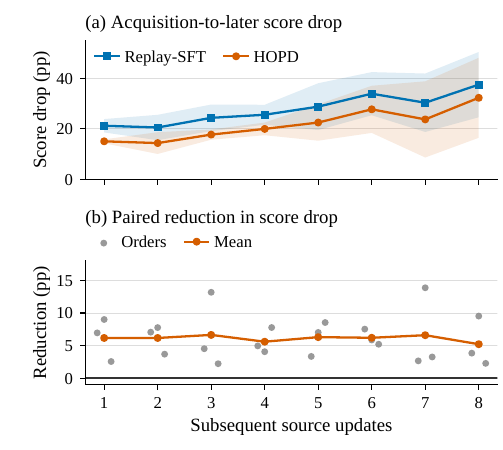}
\caption{\textbf{Retention across subsequent source updates.}
(a) Score drop from each source's completed acquisition stage, averaged over all eligible sources and seven tasks. Shading denotes source-order SD.
(b) Paired reduction in drop with HOPD. Gray dots show individual orders and the orange line their mean. HOPD reduces mean forgetting at every tested horizon in every order.}
\label{fig:retentionhorizon}
\end{figure}

%% file: tables/mechanism_controls.tex
\begin{table*}[t]
\centering
\caption{\textbf{Local consolidation and full-sequence ablations.} Stage-wise changes use shared adaptation endpoints, while overall MFN evaluates each variant's complete training trajectory. HOPD achieves the strongest recovery at both boundaries and the highest MFN.}
\label{tab:controls}
\small
\setlength{\tabcolsep}{3pt}
\begin{tabular}{llllccccc}
\toprule
Variant & Supervision & Prefix & View (T/S) & \multicolumn{2}{c}{Stage 4: HRSID} & \multicolumn{2}{c}{Stage 9: SIVED} & Full 3-sequence\\\cmidrule(lr){5-6}\cmidrule(lr){7-8}\cmidrule(lr){9-9}
 & & & & Old $\Delta\uparrow$ & Current $\Delta\uparrow$ & Old $\Delta\uparrow$ & Current $\Delta\uparrow$ & Overall MFN$\uparrow$ \\
\midrule
No consolidation & None & -- & -- & +0.00 & +0.00 & +0.00 & +0.00 & 33.09 \\
SFT consolidation & CE & GT & --/G & +1.11 & -1.08 & +21.19 & -1.11 & 49.81 \\
w/o historical gamma & History & Student & O/O & +3.30 & -0.80 & +20.54 & $\mathbf{-0.24}$ & 51.23 \\
w/ previous-stage teacher & Previous & Student & O/G & +2.57 & -1.79 & +21.26 & -0.27 & 50.56 \\
Symmetric gamma views & History & Student & G/G & +3.52 & -0.80 & +21.39 & -0.98 & 52.98 \\
GT-prefix KD & History & GT & O/G & +3.21 & -0.98 & +19.43 & -1.17 & 52.56 \\
\midrule
\textbf{HOPD} & History & Student & O/G & $\mathbf{+3.64}$ & $\mathbf{-0.43}$ & $\mathbf{+21.66}$ & -0.33 & $\mathbf{54.51}$ \\
\bottomrule
\end{tabular}
\vspace{2pt}
\begin{minipage}{\textwidth}
\scriptsize Changes are B-end minus common A-end, in pp. Old-source scores average seven tasks and all previously seen sources equally. O/G: original teacher, gamma student on historical data. G/G: identical gamma views for teacher and student. History: source-acquisition teachers. Previous: the preceding consolidated student for old sources only. Overall MFN comes from full-sequence training, not these local branches. 
\end{minipage}
\end{table*}

%% file: fig/current_intrusion.tex
\begin{figure}[t]
\centering
\includegraphics[width=\linewidth]{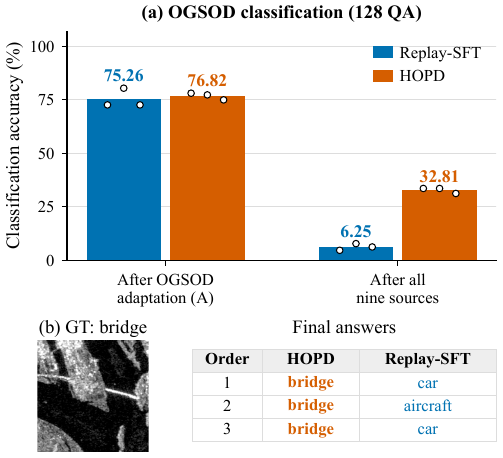}
\caption{\textbf{Category retention on OGSOD.} (a) Accuracy on 128 fixed QA at acquisition and sequence end. Bars show three-order means and dots show individual orders. (b) Final answers for a bridge correctly recognized by both methods at acquisition. HOPD better preserves earlier recognition as source priors change.}
\label{fig:currentintrusion}
\end{figure}

%% file: sec/6_conclusion.tex
\section{Limitations and Discussion}
\label{sec:limitations}
HOPD keeps deployment to one adapter, but its training-time archive grows with the number of sources.
For longer streams, selective archive compression could reduce this maintenance overhead.
The current visual augmentation operates on rendered intensity images, rather than modeling calibrated radar acquisition changes.
Extending historical consolidation to native complex-valued or polarimetric observations is a further direction.

\section{Conclusion}
\label{sec:conclusion}
We formulate source-incremental continual SAR VQA and introduce \dataset to study continual SAR interpretation as visual sources change while tasks remain fixed.
We introduce HOPD, which turns source-acquisition adapters into persistent teachers, combining student-prefix distillation and coordinate-preserving replay views to retain historical knowledge in one adapter.
Experiments show the best MFN, MAA, and BWT, alongside the best results on all four SAR behavioral diagnostics among the compared methods.
With one retained image per source, HOPD matches Replay-SFT's final performance with four.
These findings establish adaptation history as reusable supervision for continual SAR interpretation under limited image memory.

%% file: sec/S_supplement.tex
\section{Data Protocol and Scoring}
We establish \dataset from SARFact~\cite{wang2026sarfact}, a recent high-quality SAR question-answering dataset by merging and post-processing several existing datasets~\cite{zhang2025rsar,li2024sardet100k,wei2020hrsid,wang2026m4}. We only use the nine constituent datasets: OGSOD~\cite{wang2023ogsod}, Shipdataset~\cite{wang2019shipdataset}, SSDD~\cite{zhang2021ssdd}, HRSID~\cite{wei2020hrsid}, SAR-AIRcraft~\cite{wang2023saraircraft}, MSAR~\cite{xia2022msar}, AIR\_SARShip~\cite{sun2019airsarship}, SADD~\cite{zhang2022sadd}, and SIVED~\cite{lin2023sived}.
\dataset contains 356,624 training QA and 5,988 evaluation QA over 2,953 images.
Each source supports the same seven tasks: captioning, counting, presence, classification, detection, referring expressions, and spatial relations.
Public collection metadata are summarized in SARDet-100K~\cite{li2024sardet100k}, which lists satellite, airborne, and synthetic sources. 
Each of its 63 source--task evaluation cells contains 20--128 QA. The statistics are illustrated in Table~\ref{tab:currentdata}. The bounds count \emph{QA pairs}.

\input{tables/current_dataset_counts}

\begin{table*}[t]
\centering
\caption{\textbf{Primary task scores and accepted output semantics.} Scores are first computed within a source--task cell. }
\label{tab:taskprotocol}
\small
\begin{tabularx}{\textwidth}{l l Y}
\toprule
Task & Primary score & Output interpretation\\
\midrule
Caption & Core Fact F1 & Deterministic category-presence and explicit scene-count claims scored against scoped object annotations.\\
Count & Accuracy & Exact integer after unambiguous numerical parsing. \\
Presence & Accuracy & Boolean answer with explicit yes/no and true/false aliases.\\
Classification & Exact-set accuracy & Canonical category set. \\
Detection & Class-aware GRIT & One-to-one category-compatible box matching. Matched IoU sum normalized by predicted and reference object counts.\\
Referring expression & Accuracy@0.5 & One target box with IoU $\geq0.5$.\\
Spatial relation & Accuracy & Exact canonical direction among eight compass relations.\\
\bottomrule
\end{tabularx}
\end{table*}

\paragraph{Caption factual scope.}
Let $H$ be the number of extracted in-scope claims and $T$ the number supported by the annotations.
Let $G$ be the number of positive reference facts and $M$ the number matched by the predicted claims.
The score uses $P=T/H$, $R=M/G$, and $F_1=2PR/(P+R)$, with zero for an empty precision or recall denominator and for a zero harmonic-mean denominator.
For every annotated positive category, the reference facts comprise presence and its object count.
The evaluation scope includes annotated positive categories and explicitly certified absent categories. 
Extraction is independent of the reference caption and of the reference counts.
We use a conservative deterministic extractor consistent with the stated fact scope.
For a reference with two ships, ``two ships'' matches presence and count ($F_1=1$), ``ships are visible'' matches presence only ($F_1=2/3$), and ``three ships'' adds an incorrect count ($F_1=1/2$).
The extractor implements this fact scope locally, with its category aliases and parsing rules included in the evaluation code.

\paragraph{Conservative semantic normalization.}
The evaluator normalizes Unicode, capitalization, punctuation, and accepted answer wrappers, then applies explicit task-specific aliases.
For structured answers, conflicting alternatives and unresolved negation are rejected. Unknown categories in category-set and box outputs remain errors or false positives.
This supports linguistic variation without crediting a semantically different answer merely because its text is similar.

\paragraph{Detection and coordinate conventions.}
The default annotation convention is continuous $xyxy$ coordinates expressed on a 0--1000 scale and normalized for overlap computation, with explicit per-row convention overrides when supplied.
The same convention is applied to predictions and references. No hidden per-example coordinate calibration uses the answer.
For current GRIT scoring~\cite{gupta2022grit}, incompatible categories receive zero pairwise overlap, and one-to-one assignment~\cite{kuhn1955hungarian} maximizes total IoU.
If $N_p$ and $N_g$ count predicted and reference objects and $N_+$ counts positive-IoU assignments, the score is the matched IoU sum divided by $N_p+N_g-N_+$.
There is no IoU-$0.5$ cutoff or confidence ranking in this metric.
Valid empty prediction and reference sets score one. Missing responses and wholly unparseable responses score zero.

\section{Implementation Details}
\begin{algorithm}[t]
\caption{Recursive HOPD with historical cross-view transfer}
\label{alg:historical}
\begin{algorithmic}[1]
\Require Initial adapter $\theta_0$, sources $\mathcal D_{1:T}$, image budget $n$ per source, $K$ tasks
\State $\memory\gets\varnothing$, $\Phi\gets\varnothing$
\For{$t=1,\ldots,T$}
 \State $\theta_t^A\gets\operatorname{SFT}(\theta_{t-1},\mathcal D_t)$
 \State $\phi_t\gets\operatorname{freeze}(\theta_t^A)$ and add to $\Phi$
 \State $\theta_t\gets\theta_t^A$
 \If{$t>1$}
  \For{each consolidation update}
   \State Draw a QA minibatch: uniform source, then uniform task
   \State Set $I^s_{i,t}=G_\gamma(I)$ if $i<t$, otherwise $I$
   \State Sample student answers on $I^s_{i,t}$
   \State Query $\phi_i$ on $I$ and the same answer prefixes
   \State Update $\theta_t$ with full-vocabulary $\KL(p_{\phi_i}\|p_{\theta_t})$
  \EndFor
 \EndIf
 \State Commit $n$ images and their $nK$ task QA to $\memory_t$
 \State Evaluate $\theta_t$ on original images from sources $1{:}t$
\EndFor
\State \Return $\theta_T$ for inference
\end{algorithmic}
\end{algorithm}

\paragraph{Backbone and adapters.}
By default, we use Qwen3.5-4B~\cite{qwen2026qwen35} as the baseline model. HOPD uses LoRA adapters~\cite{hu2022lora} to update the model.
LoRA modifies 346 linear maps: 96 in the vision encoder, two in the merger, and 248 in the language model.
Rank and $\alpha$ are both 8 (multiplier $\alpha/r=1$), adapter dropout is zero, and the backbone, embeddings, normalizations, and vocabulary head are frozen.
The model receives an image, the question, and a task-specific output-format hint. Source provenance and reference-answer metadata are excluded from the prompt.
Thinking output is disabled.

\paragraph{Declared source orders.}
Let source indices $1$--$9$ denote OGSOD~\cite{wang2023ogsod}, Shipdataset~\cite{wang2019shipdataset}, SSDD~\cite{zhang2021ssdd}, HRSID~\cite{wei2020hrsid}, SAR-AIRcraft~\cite{wang2023saraircraft}, MSAR~\cite{xia2022msar}, AIR\_SARShip~\cite{sun2019airsarship}, SADD~\cite{zhang2022sadd}, and SIVED~\cite{lin2023sived}, respectively.
The configuration specifies $(1,2,3,4,5,6,7,8,9)$, $(6,7,9,4,1,3,5,2,8)$, and $(5,1,8,7,6,2,3,4,9)$, which are randomly generated.

Three complete source orders are measured for the main method and baselines. Local controls branch at stages four and nine of the first order.

\paragraph{Optimization and normalization.}
AdamW~\cite{loshchilov2019adamw} uses $(\beta_1,\beta_2)=(0.9,0.999)$, $\epsilon=10^{-8}$, weight decay 0, and gradient-norm clipping at 1.
Each phase has its own optimizer/schedule initialization.
The effective batch is eight QA pairs. Main full runs use four replicated processes. Local controls use two L40 GPUs, microbatch one and native CE batching disabled for every arm.
During B, the main configuration uses one rollout per selected QA and microbatch limit 1.
The loss reduction is token mean, then rollout mean, then QA mean. 

\paragraph{Replay-SFT update budget.}
Replay-SFT in the main comparison interleaves current-source and historical-memory QA during 1,024 supervised updates per source.
HOPD performs 1,024 Stage-A updates and an additional 128 Stage-B updates at D2--D9.
We also tested appending 128 supervised consolidation updates to Replay-SFT after its replay-based training phase.
This extension produced pronounced overfitting and a marked decline in evaluation performance.
We therefore report the stronger Replay-SFT configuration without the extra consolidation phase.
The comparison uses the same image-memory budget but method-specific update schedules, since matching HOPD's update count degraded this baseline in our experiments.

\paragraph{End tokens and numerical details.}
The first model/chat stop token is retained and contributes to the loss. 
A first-token EOS response is a valid one-token rollout even if its decoded text is empty.
Ground-truth sequences are never truncated. Exceeding the separate GT length cap raises an error.

Teacher targets are detached. The main full-vocabulary forward KL is evaluated in FP32 with gradients through student log probabilities. Reverse KL is a labeled sensitivity variant.
Vocabulary projections are processed in answer-position chunks of 64 to limit temporary logits memory without truncating the vocabulary.

\paragraph{Teacher and exemplar storage.}
The main experiments instantiate the memory budget with $n=1$ and $K=7$: one image and seven native task QA per source, totaling nine images and 63 QA after nine sources.

The current source is sampled from full TRAIN during stage B and enters memory only after B. 
Historical source and task selection is uniform, independent of pool size.
Source-specific A-end teachers are saved before B and remain frozen. 
The final student uses one adapter at deployment.

\paragraph{Gamma recipe and activation.}
An RGB lookup table applies the same monotone gamma transform to all channels before native image preprocessing. Image dimensions and the model's patch-grid layout are preserved.
Each historical student rollout and its gradient update use the same transformed pixels and generated token sequence.
The frozen teacher sees the original image and the identical integer prefix, with native precision and no extra gamma-specific autocast.
Current-source queries, adaptation, and inference retain original images.
The symmetric control substitutes those exact transformed pixels for the historical teacher as well. 
The main configuration uses source-history teachers, student rollouts, full-vocabulary forward KL and one rollout per QA. Dedicated control branches explicitly record their changed objectives and prefix sources.

\section{Additional Metric Definitions}
\label{sec:additionalmetrics}
\subsection{Conventional continual-learning summaries}
Let $R^{(k)}_{t,i}$ denote the primary score for task $k$, with $R_{t,i}=K^{-1}\sum_k R^{(k)}_{t,i}$.
All seven tasks are supported for every source in this release, so task-first and source-first macro averaging commute.
MAA weights completed stages equally.
Adapting peak-referenced forgetting~\cite{chaudhry2018rwalk}, we additionally define
\begin{equation}
 F=\frac{1}{T-1}\sum_{i<T}
 \left[\max_{i\leq t\leq T} R_{t,i}-R_{T,i}\right].
\end{equation}
Including the final stage in the maximum makes $F$ nonnegative.
A per-task implementation should compute each task's peak separately before averaging if that is the reported convention. 
The main table uses acquisition-to-final drop. Pre-B and post-B reference points are kept separate.
Worst-source final performance is $\min_i R_{T,i}$.

\input{sec/S_domain_diagnostics}

\section{Additional Results}

\paragraph{Where final retention improves.}
\Cref{fig:taskdeltas} resolves the aggregate gain into source--task cells.
The largest improvements occur in classification, including +31.7 percentage points (pp) on AIR\_SARShip and +26.6 pp on OGSOD.
Detection improves on every source, while referring-expression scores improve or remain unchanged, showing that the benefit extends to spatial predictions.
Averaging across sources, HOPD improves all seven task scores, linking the overall gain to both category recognition and spatial interpretation.

\paragraph{Effect of consolidation on the arriving source.}
\Cref{fig:currentcost} compares each arriving source immediately before and after phase B.
Positive values indicate improved current-source performance, and negative values indicate a loss after consolidation.
Across stages two through nine and the three orders, the mean change is $-0.57$ pp for HOPD and $-1.15$ pp for its reverse-KL variant.
The forward-KL recipe retains more of the newly acquired performance on average while consolidating historical knowledge.

\paragraph{Forgetting from the Stage-A endpoint.}
To check whether the retention advantage depends on the acquisition reference, we replace each source's completed-stage score with its score immediately after Stage A, before consolidation.
For order $o$, let $A_i^{(o)}$ denote this task-macro score for source $i$ and $R_{i+k,i}^{(o)}$ its score after $k$ subsequent source updates.
We compute
\begin{equation}
 F_o^A(k)=\frac{1}{T-k}\sum_{i=1}^{T-k}
 \left(A_i^{(o)}-R_{i+k,i}^{(o)}\right),
\end{equation}
where $T=9$ and scores average the seven tasks.
The checkpoints and evaluation examples are unchanged from the main retention-horizon analysis, giving $3(9-k)$ matched source--stage pairs per horizon.
\Cref{tab:stageareference} reports the resulting drops and paired reductions.
HOPD reduces forgetting at every horizon in every order, showing that its advantage persists when measured from the pre-consolidation acquisition state.
\input{tables/stage_a_retention}

%% file: tables/current_dataset_counts.tex
\begin{table}[t]
\centering
\caption{SARVQA-CL dataset statistics.}
\label{tab:currentdata}
\small
\setlength{\tabcolsep}{4pt}
\begin{tabular}{lrr}
\toprule
Source & Train QA & Eval QA \\
\midrule
OGSOD & 41,301 & 891 \\
Shipdataset & 106,167 & 830 \\
SSDD & 5,221 & 468 \\
HRSID & 26,500 & 725 \\
SAR-AIRcraft & 53,456 & 896 \\
MSAR & 110,226 & 896 \\
AIR\_SARShip & 3,949 & 197 \\
SADD & 3,937 & 497 \\
SIVED & 5,867 & 588 \\
\midrule
Total & 356,624 & 5,988 \\
\bottomrule
\end{tabular}
\end{table}

%% file: sec/S_domain_diagnostics.tex
\subsection{Domain-transition diagnostics}
\label{sec:domaindiagnostics}
The primary scores quantify task performance and its evolution
Source-conditioned answer distributions motivate the first question: after learning a new SAR domain, do previously correct old-domain answers acquire its dominant category?
The shared count and detection tasks motivate the second: does an image retain both a correct object count and a complete localization of those objects?
We introduce dominant-class intrusion rate (DCIR) and joint fact retention (JFR) as complementary diagnostics of these two behaviors.
They supplement MFT, MFN, MAA, and BWT without changing the seven-task overall score, its task metrics, or the training protocol.

\paragraph{Stage boundary.}
Let $t$ index domains in the training sequence and $i<t$ an old domain.
The stage-$t$ model is evaluated after completing that domain: after phase B when consolidation is present, and after phase A otherwise.
Each transition compares the stage-$(t-1)$ and stage-$t$ models on fixed old-domain evaluation examples.

\input{tables/diagnostic_coverage}

\subsection{Dominant-class intrusion rate}
\label{sec:dcir}
\paragraph{Incoming-domain category.}
We identify the dominant category using only the incoming domain's training classification annotations.
For each canonical category $c$, let $n_t(c)$ be the number of training QA whose reference category set contains $c$.
A multilabel QA contributes once to every category in its set.
The dominant category $c_t$ is the unique maximizer of $n_t(c)$.
Ties, absent categories, or missing training annotations make this diagnostic unavailable. Evaluation labels are never used to select $c_t$.

\paragraph{Conditional intrusion and forgetting.}
For an old-domain classification QA $q$, denote its reference set by $Y(q)$ and its stage-$t$ predicted set by $\widehat{Y}_t(q)$.
Let $a_t(q)=1$ precisely when the prediction is valid and matches $Y(q)$ exactly, and $a_t(q)=0$ otherwise.
The eligible set is
\begin{equation}
 \mathcal{E}_{t,i}=\{q\in\mathcal{Q}^{\mathrm{cls}}_i:
 a_{t-1}(q)=1,\ c_t\notin Y(q)\},
 \label{eq:dcir-eligible}
\end{equation}
where $\mathcal{Q}^{\mathrm{cls}}_i$ is the fixed classification evaluation set for domain $i$.
Define $h_t(q)=1$ if the current prediction is valid and contains $c_t$, and zero otherwise.
Then
\begin{align}
 \mathrm{DCIR}_{t,i}
 &=\frac{\sum_{q\in\mathcal{E}_{t,i}}h_t(q)}{|\mathcal{E}_{t,i}|},
 \label{eq:dcir}\\
 \mathrm{NFR}_{t,i}
 &=\frac{\sum_{q\in\mathcal{E}_{t,i}}[1-a_t(q)]}{|\mathcal{E}_{t,i}|}.
 \label{eq:nfr}
\end{align}
Lower values are better for both rates.
The companion NFR adapts negative-flip analysis~\cite{yan2021positive} to the \emph{same} eligible set, counting newly incorrect answers, missing responses, and parsing failures.
DCIR counts only errors that introduce the incoming dominant category, so $0\leq\mathrm{DCIR}_{t,i}\leq\mathrm{NFR}_{t,i}\leq1$ whenever both rates are defined.
A malformed answer can increase NFR without increasing DCIR.

\paragraph{Interpretation.}
We report DCIR together with NFR and incoming-domain classification accuracy.
This distinguishes low directional intrusion from low overall forgetting and helps assess whether low intrusion comes at the cost of learning the new domain.
DCIR diagnoses a specific direction of output interference.
It measures changes in predicted categories, while visual dependence would require interventions on the input image.

\subsection{Joint fact retention}
\label{sec:jfr}
\paragraph{Constructing fixed fact pairs.}
We form a set $\mathcal{F}_i$ of count--detection pairs from the evaluated QA in domain $i$ using the key (source, split, image, canonical query category, full-image scope).
Each key must identify exactly one count QA and one detection QA concerning the same category over the entire image.
Full-image scope is established by strict matching to the native question template and its query category, or by explicit full-image scope metadata. Sharing an image alone is insufficient.
The reference count must equal the number of reference boxes for that category.
Candidates with missing pairing metadata, other reference categories, ambiguous scope, duplicate pairings, or inconsistent references are excluded with recorded reasons.
Pair construction depends on annotations, and the resulting pairs remain fixed across stages and methods evaluated on the same QA and pairing metadata.

\input{tables/main_results_full}

\begin{figure*}[t]
\centering
\includegraphics[width=\textwidth]{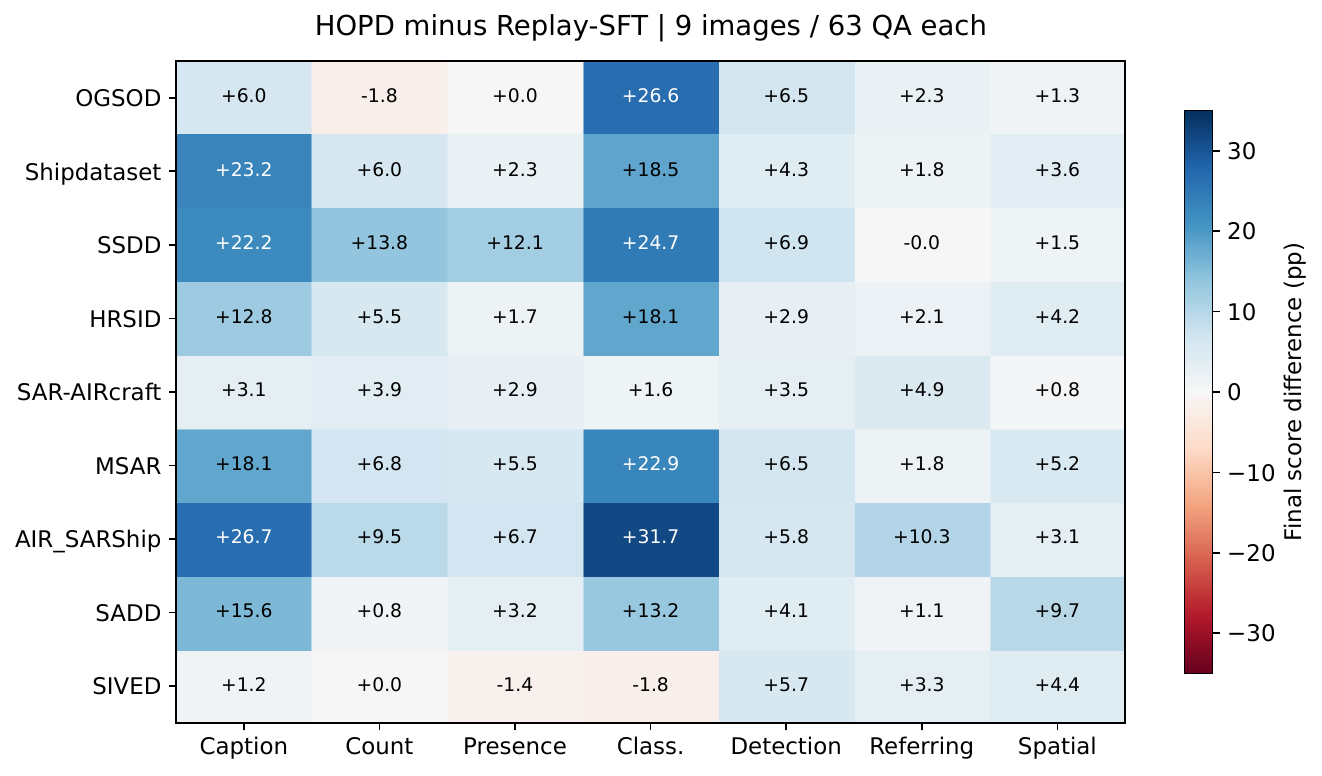}
\caption{\textbf{Final source--task gains over Replay-SFT.} HOPD minus Replay-SFT, averaged across three orders. HOPD improves every task on average, with detection gains on all sources.}
\label{fig:taskdeltas}
\end{figure*}
\begin{figure*}[t]
\centering
\includegraphics[width=\textwidth]{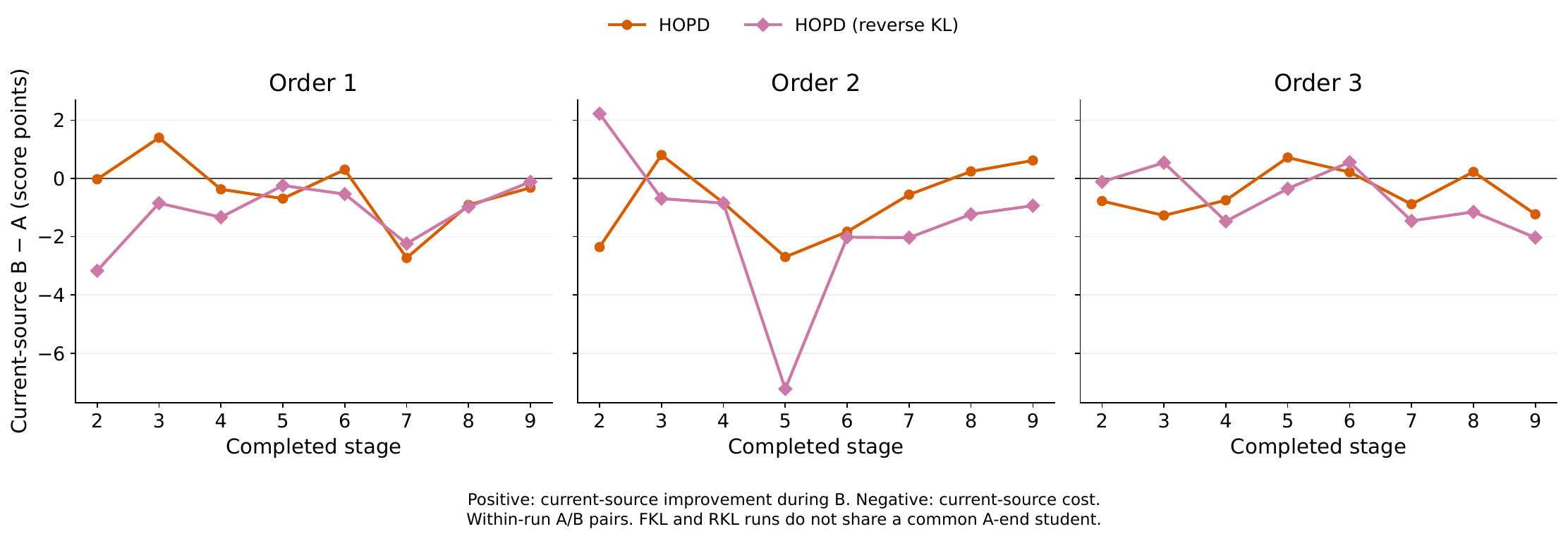}
\caption{\textbf{Current-source performance through consolidation.} Seven-task B-end minus A-end changes at D2--D9 along each recipe's full trajectory. Forward KL better preserves acquired current-source performance on average.}
\label{fig:currentcost}
\end{figure*}

\paragraph{Joint correctness.}
For a fact pair $f$, let $J_t(f)=1$ if and only if the stage-$t$ count is correct and the detection response is completely parseable, has valid box geometry and the correct categories, and admits a complete one-to-one matching between predicted and reference boxes with IoU $\geq0.5$ for every match.
No extra prediction or missed reference is allowed.
We use maximum-cardinality bipartite matching on category-correct box pairs at this threshold and require a matching covering both sets.
An empty reference is jointly correct when the count is zero and the detection response is a valid, complete empty list.
Otherwise $J_t(f)=0$, including for recorded missing responses or invalid outputs.
This criterion is stricter than the primary GRIT scorer, which can retain complete objects before a truncated trailing object: JFR requires the entire detection response to be valid.
For example, a correct count of three paired with only two correctly localized objects receives no joint-correctness credit.

\paragraph{Retention and current accuracy.}
We measure retention among the facts that were jointly correct immediately before the transition:
\begin{equation}
 \mathrm{JFR}_{t,i}=
 \frac{\sum_{f\in\mathcal{F}_i}J_{t-1}(f)J_t(f)}
 {\sum_{f\in\mathcal{F}_i}J_{t-1}(f)}.
 \label{eq:jfr}
\end{equation}
Its companion joint fact accuracy (JFA) evaluates all valid fixed pairs in the same old domain:
\begin{equation}
 \mathrm{JFA}_{t,i}=\frac{1}{|\mathcal{F}_i|}
 \sum_{f\in\mathcal{F}_i}J_t(f).
 \label{eq:jfa}
\end{equation}
Higher values are better for both metrics.
JFA makes the absolute level of joint correctness visible when JFR conditions on a small previously correct subset.
Count--detection agreement alone earns no credit when the two answers are wrong.

\paragraph{Reporting convention.}
All four diagnostics are macro-averaged over valid old sources within each transition, then over valid transitions.
Sequence JFA uses this transition average, rather than final-stage accuracy over all seen sources.
We report the equal-weight mean and population standard deviation over complete valid orders.
DCIR/NFR eligibility and the JFR denominator depend on each method's previously correct predictions, so conditional sets can differ across methods.
\Cref{tab:diagnosticcoverage} reports available coverage as repeated eligible occurrences.

%% file: tables/diagnostic_coverage.tex
\begin{table*}[t]
\centering
\caption{\textbf{Coverage of behavioral diagnostics.} Counts sum eligible occurrences over old sources and transitions. Both methods provide diagnostic coverage across all eight transitions in every order.}
\label{tab:diagnosticcoverage}
\small
\setlength{\tabcolsep}{4pt}
\begin{tabular}{llccccc}
\toprule
Method & Order & DCIR/NFR QA & JFR pairs & JFA pairs & Empty DCIR cells & Empty JFR cells\\
\midrule
HOPD & 1 & 1677 & 435 & 1195 & 10 & 9 \\
HOPD & 2 & 1547 & 384 & 1472 & 7 & 4 \\
HOPD & 3 & 1233 & 270 & 1180 & 9 & 10 \\
Replay-SFT & 1 & 1324 & 387 & 1195 & 10 & 9 \\
Replay-SFT & 2 & 1316 & 339 & 1472 & 8 & 5 \\
Replay-SFT & 3 & 1101 & 246 & 1180 & 9 & 11 \\
\bottomrule
\end{tabular}
\vspace{2pt}
\begin{minipage}{\textwidth}
\scriptsize JFA counts use the fixed annotation-valid pair set. JFR counts use previously jointly correct pairs. 
\end{minipage}
\end{table*}

%% file: tables/main_results_full.tex
\begin{table*}[t]
\centering
\caption{\textbf{Main results with variation across source orders.} Mean $\pm$ population SD over three orders. HOPD has the highest mean final performance and the lowest mean forgetting drop.}
\label{tab:mainfull}
\scriptsize
\setlength{\tabcolsep}{3pt}
\begin{tabular}{lrrrrrrrr}
\toprule
Method & MFN$\uparrow$ & MAA$\uparrow$ & BWT$\downarrow$ & MFT$\uparrow$ & DCIR$\downarrow$ & NFR$\downarrow$ & JFR$\uparrow$ & JFA$\uparrow$ \\
\midrule
Zero-shot & 20.50 & \NA & \NA & \NA & \NA & \NA & \NA & \NA \\
Sequential LoRA & $33.09\pm 3.32$ & $44.83\pm 2.57$ & $43.47\pm 3.16$ & $71.73\pm 0.59$ & $84.98\pm 13.96$ & $91.06\pm 9.77$ & $20.68\pm 9.04$ & $9.31\pm 2.88$ \\
Replay-SFT & $47.06\pm 0.38$ & $50.23\pm 2.79$ & $26.99\pm 0.36$ & $71.05\pm 0.43$ & $49.98\pm 6.84$ & $55.29\pm 3.04$ & $47.98\pm 4.47$ & $15.43\pm 4.09$ \\
EWC-LoRA & $33.74\pm 1.69$ & $45.22\pm 2.55$ & $42.36\pm 1.96$ & $71.39\pm 0.28$ & $82.97\pm 9.11$ & $90.00\pm 6.42$ & $22.49\pm 11.01$ & $9.90\pm 2.90$ \\
LwF & $33.01\pm 1.75$ & $45.73\pm 2.41$ & $42.96\pm 2.00$ & $71.19\pm 0.16$ & $89.36\pm 4.00$ & $94.92\pm 1.55$ & $29.60\pm 4.34$ & $9.73\pm 2.97$ \\
O-LoRA & $35.95\pm 3.91$ & $46.01\pm 2.62$ & $37.38\pm 4.80$ & $69.18\pm 0.36$ & $77.88\pm 5.15$ & $83.39\pm 4.17$ & $36.51\pm 4.23$ & $12.10\pm 3.70$ \\
RegLoRA & $35.96\pm 3.48$ & $46.43\pm 2.51$ & $39.02\pm 3.76$ & $70.65\pm 0.15$ & $81.30\pm 10.49$ & $89.23\pm 6.91$ & $38.50\pm 7.87$ & $11.82\pm 2.58$ \\
SDFT & \reviewmark{$18.66\pm 4.44$} & $30.11\pm 1.82$ & $22.18\pm 7.75$ & $38.38\pm 2.77$ & $46.44\pm 8.10$ & $59.71\pm 18.45$ & $0.00\pm 0.00$ & $0.00\pm 0.00$ \\
MIR & $50.72\pm 2.22$ & $51.44\pm 1.96$ & $19.35\pm 2.42$ & $67.92\pm 0.07$ & $38.36\pm 3.70$ & $46.21\pm 0.39$ & $45.35\pm 2.48$ & $14.31\pm 2.92$ \\
MR-LoRA & $51.55\pm 1.29$ & $52.26\pm 2.20$ & $21.57\pm 3.97$ & $70.72\pm 2.45$ & $35.36\pm 2.33$ & $43.08\pm 3.06$ & $47.03\pm 0.72$ & $16.16\pm 5.47$ \\
\midrule
\textbf{HOPD} & $54.51\pm 1.31$ & $54.60\pm 2.93$ & $18.58\pm 1.18$ & $71.03\pm 0.38$ & $32.97\pm 2.56$ & $39.23\pm 1.49$ & $56.06\pm 1.49$ & $17.79\pm 4.03$ \\
\bottomrule
\end{tabular}  
\end{table*}

%% file: tables/stage_a_retention.tex
\begin{table}[t]
\centering
\caption{\textbf{Forgetting from the Stage-A endpoint.} Mean $\pm$ population SD across three source orders, in pp. Reduction is Replay-SFT minus HOPD, computed within each order. HOPD reduces forgetting at every horizon in all three orders.}
\label{tab:stageareference}
\small
\setlength{\tabcolsep}{3pt}
\begin{tabular}{crrr}
\toprule
$k$ & Replay-SFT $\downarrow$ & HOPD $\downarrow$ & Reduction $\uparrow$ \\
\midrule
1 & $21.23\pm2.65$ & $15.59\pm0.62$ & $5.64\pm2.48$ \\
2 & $20.53\pm5.03$ & $14.95\pm4.08$ & $5.58\pm1.62$ \\
3 & $24.36\pm5.26$ & $18.18\pm2.36$ & $6.19\pm4.27$ \\
4 & $25.58\pm3.96$ & $20.44\pm2.64$ & $5.14\pm1.76$ \\
5 & $28.80\pm9.28$ & $22.85\pm7.56$ & $5.95\pm1.77$ \\
6 & $33.93\pm8.55$ & $27.98\pm9.71$ & $5.95\pm1.48$ \\
7 & $30.34\pm11.58$ & $24.27\pm14.82$ & $6.07\pm4.68$ \\
8 & $37.52\pm12.94$ & $32.31\pm15.85$ & $5.22\pm3.11$ \\
\bottomrule
\end{tabular}
\end{table}